\documentclass{article}

\usepackage[numbers]{natbib}

\usepackage[final]{neurips_2021}

\usepackage[utf8]{inputenc} 
\usepackage[T1]{fontenc}    
\usepackage{hyperref}       
\usepackage{url}            
\usepackage{booktabs}       
\usepackage{amsfonts}       
\usepackage{nicefrac}       
\usepackage{microtype}      
\usepackage{xcolor}         

\usepackage{amsthm,amsmath,bm,xspace}
\usepackage{booktabs}
\usepackage{mathtools}
\usepackage{subcaption}
\usepackage{caption}
\usepackage{todonotes}

\title{Magic Pyramid: Accelerating Inference with \\ Early Exiting and Token Pruning}

\author{%
\begin{tabular}{cccc}
   Xuanli He$^1$  &  Iman Keivanloo$^2$ & Yi Xu$^2$ & 
   Xiang He$^2$
\end{tabular} \\
\begin{tabular}{ccc}
  \tabularnewline \textbf{Belinda Zeng}$^2$ & \textbf{Santosh Rajagopalan} $^2$  & \textbf{Trishul Chilimbi}$^2$ 
\end{tabular} \\

\begin{tabular}{c@{\hspace{1cm}}c}
  \tabularnewline $^1$Monash University, Australia & $^2$Amazon
\end{tabular}\\[.05cm]
\texttt{xuanli.he1@monash.edu, \{imankei,yxaamzn\}@amazon.com} \\
}

\begin{document}

\maketitle

\begin{abstract}
  Pre-training and then fine-tuning large language models is  commonly used  to achieve state-of-the-art performance in natural language processing (NLP) tasks. However, most pre-trained models suffer from low inference speed. Deploying such large models to applications with latency constraints is challenging. In this work, we focus on accelerating the inference via conditional computations. To achieve this, we propose a novel idea, Magic Pyramid (MP), to reduce both width-wise and depth-wise computation via token pruning and early exiting for Transformer-based models, particularly BERT. The former manages to save the computation via removing non-salient tokens, while the latter can fulfill the computation reduction by terminating the inference early before reaching the final layer, if the exiting condition is met. Our empirical studies demonstrate that compared to previous state of arts, MP is not only able to achieve a speed-adjustable inference, but also to surpass token pruning and early exiting by reducing up to 70\% giga floating point operations (GFLOPs) with less than 0.5\% accuracy drop. Token pruning and early exiting express distinctive preferences to sequences with different lengths. However, MP is capable of achieving an average of 8.06x speedup on two popular text classification tasks, regardless of the sizes of the inputs.
\end{abstract}


\def\eg{{\em e.g.,}\xspace}
\def\ie{{\em i.e.,}\xspace}
\def\versus{{\em v.s.}\xspace}
\def\cf{{\em c.f.,}\xspace}
\def\etc{{\em etc.,}\xspace}

\newcommand{\hxl}[1]{\textcolor{red}{\small\textbf{[XL:]} #1}}

\newcommand{\mohammad}[1]{{\textcolor{blue}{#1}}}
\newcommand{\reza}[1]{{\textcolor{orange}{#1}}}
\newcommand{\xuanli}[1]{{\textcolor{red}{#1}}}

\newcommand{\red}[1]{{\bf \textcolor{red}{#1}}}

\newcommand{\gal}{GAL\xspace}

\newcommand{\figleft}{{\em (Left)}}
\newcommand{\figcenter}{{\em (Center)}}
\newcommand{\figright}{{\em (Right)}}
\newcommand{\figtop}{{\em (Top)}}
\newcommand{\figbottom}{{\em (Bottom)}}
\newcommand{\captiona}{{\em (a)}}
\newcommand{\captionb}{{\em (b)}}
\newcommand{\captionc}{{\em (c)}}
\newcommand{\captiond}{{\em (d)}}

\newcommand{\newterm}[1]{{\bf #1}}

\def\figref#1{Figure~\ref{#1}}
\def\Figref#1{Figure~\ref{#1}}
\def\twofigref#1#2{figures \ref{#1} and \ref{#2}}
\def\quadfigref#1#2#3#4{figures \ref{#1}, \ref{#2}, \ref{#3} and \ref{#4}}
\def\tabref#1{Table~\ref{#1}}
\def\Tabref#1{Table~\ref{#1}}
\def\secref#1{section~\ref{#1}}
\def\Secref#1{Section~\ref{#1}}
\def\twosecrefs#1#2{sections \ref{#1} and \ref{#2}}
\def\secrefs#1#2#3{sections \ref{#1}, \ref{#2} and \ref{#3}}
\def\eqref#1{(\ref{#1})}
\def\Eqref#1{Equation~\ref{#1}}
\def\plaineqref#1{\ref{#1}}
\def\chapref#1{chapter~\ref{#1}}
\def\Chapref#1{Chapter~\ref{#1}}
\def\rangechapref#1#2{chapters\ref{#1}--\ref{#2}}
\def\algref#1{algorithm~\ref{#1}}
\def\Algref#1{Algorithm~\ref{#1}}
\def\twoalgref#1#2{Algorithms \ref{#1} and \ref{#2}}
\def\Twoalgref#1#2{Algorithms \ref{#1} and \ref{#2}}
\def\partref#1{part~\ref{#1}}
\def\Partref#1{Part~\ref{#1}}
\def\twopartref#1#2{parts \ref{#1} and \ref{#2}}

\def\ceil#1{\lceil #1 \rceil}
\def\floor#1{\lfloor #1 \rfloor}
\def\1{\bm{1}}
\newcommand{\train}{\mathcal{D}}
\newcommand{\valid}{\mathcal{D_{\mathrm{valid}}}}
\newcommand{\test}{\mathcal{D_{\mathrm{test}}}}

\def\eps{{\epsilon}}

\def\reta{{\textnormal{$\eta$}}}
\def\ra{{\textnormal{a}}}
\def\rb{{\textnormal{b}}}
\def\rc{{\textnormal{c}}}
\def\rd{{\textnormal{d}}}
\def\re{{\textnormal{e}}}
\def\rf{{\textnormal{f}}}
\def\rg{{\textnormal{g}}}
\def\rh{{\textnormal{h}}}
\def\ri{{\textnormal{i}}}
\def\rj{{\textnormal{j}}}
\def\rk{{\textnormal{k}}}
\def\rl{{\textnormal{l}}}
\def\rn{{\textnormal{n}}}
\def\ro{{\textnormal{o}}}
\def\rp{{\textnormal{p}}}
\def\rq{{\textnormal{q}}}
\def\rr{{\textnormal{r}}}
\def\rs{{\textnormal{s}}}
\def\rt{{\textnormal{t}}}
\def\ru{{\textnormal{u}}}
\def\rv{{\textnormal{v}}}
\def\rw{{\textnormal{w}}}
\def\rx{{\textnormal{x}}}
\def\ry{{\textnormal{y}}}
\def\rz{{\textnormal{z}}}

\def\rvepsilon{{\mathbf{\epsilon}}}
\def\rvtheta{{\mathbf{\theta}}}
\def\rva{{\mathbf{a}}}
\def\rvb{{\mathbf{b}}}
\def\rvc{{\mathbf{c}}}
\def\rvd{{\mathbf{d}}}
\def\rve{{\mathbf{e}}}
\def\rvf{{\mathbf{f}}}
\def\rvg{{\mathbf{g}}}
\def\rvh{{\mathbf{h}}}
\def\rvu{{\mathbf{i}}}
\def\rvj{{\mathbf{j}}}
\def\rvk{{\mathbf{k}}}
\def\rvl{{\mathbf{l}}}
\def\rvm{{\mathbf{m}}}
\def\rvn{{\mathbf{n}}}
\def\rvo{{\mathbf{o}}}
\def\rvp{{\mathbf{p}}}
\def\rvq{{\mathbf{q}}}
\def\rvr{{\mathbf{r}}}
\def\rvs{{\mathbf{s}}}
\def\rvt{{\mathbf{t}}}
\def\rvu{{\mathbf{u}}}
\def\rvv{{\mathbf{v}}}
\def\rvw{{\mathbf{w}}}
\def\rvx{{\mathbf{x}}}
\def\rvy{{\mathbf{y}}}
\def\rvz{{\mathbf{z}}}

\def\erva{{\textnormal{a}}}
\def\ervb{{\textnormal{b}}}
\def\ervc{{\textnormal{c}}}
\def\ervd{{\textnormal{d}}}
\def\erve{{\textnormal{e}}}
\def\ervf{{\textnormal{f}}}
\def\ervg{{\textnormal{g}}}
\def\ervh{{\textnormal{h}}}
\def\ervi{{\textnormal{i}}}
\def\ervj{{\textnormal{j}}}
\def\ervk{{\textnormal{k}}}
\def\ervl{{\textnormal{l}}}
\def\ervm{{\textnormal{m}}}
\def\ervn{{\textnormal{n}}}
\def\ervo{{\textnormal{o}}}
\def\ervp{{\textnormal{p}}}
\def\ervq{{\textnormal{q}}}
\def\ervr{{\textnormal{r}}}
\def\ervs{{\textnormal{s}}}
\def\ervt{{\textnormal{t}}}
\def\ervu{{\textnormal{u}}}
\def\ervv{{\textnormal{v}}}
\def\ervw{{\textnormal{w}}}
\def\ervx{{\textnormal{x}}}
\def\ervy{{\textnormal{y}}}
\def\ervz{{\textnormal{z}}}

\def\rmA{{\mathbf{A}}}
\def\rmB{{\mathbf{B}}}
\def\rmC{{\mathbf{C}}}
\def\rmD{{\mathbf{D}}}
\def\rmE{{\mathbf{E}}}
\def\rmF{{\mathbf{F}}}
\def\rmG{{\mathbf{G}}}
\def\rmH{{\mathbf{H}}}
\def\rmI{{\mathbf{I}}}
\def\rmJ{{\mathbf{J}}}
\def\rmK{{\mathbf{K}}}
\def\rmL{{\mathbf{L}}}
\def\rmM{{\mathbf{M}}}
\def\rmN{{\mathbf{N}}}
\def\rmO{{\mathbf{O}}}
\def\rmP{{\mathbf{P}}}
\def\rmQ{{\mathbf{Q}}}
\def\rmR{{\mathbf{R}}}
\def\rmS{{\mathbf{S}}}
\def\rmT{{\mathbf{T}}}
\def\rmU{{\mathbf{U}}}
\def\rmV{{\mathbf{V}}}
\def\rmW{{\mathbf{W}}}
\def\rmX{{\mathbf{X}}}
\def\rmY{{\mathbf{Y}}}
\def\rmZ{{\mathbf{Z}}}

\def\ermA{{\textnormal{A}}}
\def\ermB{{\textnormal{B}}}
\def\ermC{{\textnormal{C}}}
\def\ermD{{\textnormal{D}}}
\def\ermE{{\textnormal{E}}}
\def\ermF{{\textnormal{F}}}
\def\ermG{{\textnormal{G}}}
\def\ermH{{\textnormal{H}}}
\def\ermI{{\textnormal{I}}}
\def\ermJ{{\textnormal{J}}}
\def\ermK{{\textnormal{K}}}
\def\ermL{{\textnormal{L}}}
\def\ermM{{\textnormal{M}}}
\def\ermN{{\textnormal{N}}}
\def\ermO{{\textnormal{O}}}
\def\ermP{{\textnormal{P}}}
\def\ermQ{{\textnormal{Q}}}
\def\ermR{{\textnormal{R}}}
\def\ermS{{\textnormal{S}}}
\def\ermT{{\textnormal{T}}}
\def\ermU{{\textnormal{U}}}
\def\ermV{{\textnormal{V}}}
\def\ermW{{\textnormal{W}}}
\def\ermX{{\textnormal{X}}}
\def\ermY{{\textnormal{Y}}}
\def\ermZ{{\textnormal{Z}}}

\def\vzero{{\bm{0}}}
\def\vone{{\bm{1}}}
\def\vmu{{\bm{\mu}}}
\def\vtheta{{\bm{\theta}}}
\def\va{{\bm{a}}}
\def\vb{{\bm{b}}}
\def\vc{{\bm{c}}}
\def\vd{{\bm{d}}}
\def\ve{{\bm{e}}}
\def\vf{{\bm{f}}}
\def\vg{{\bm{g}}}
\def\vh{{\bm{h}}}
\def\vi{{\bm{i}}}
\def\vj{{\bm{j}}}
\def\vk{{\bm{k}}}
\def\vl{{\bm{l}}}
\def\vm{{\bm{m}}}
\def\vn{{\bm{n}}}
\def\vo{{\bm{o}}}
\def\vp{{\bm{p}}}
\def\vq{{\bm{q}}}
\def\vr{{\bm{r}}}
\def\vs{{\bm{s}}}
\def\vt{{\bm{t}}}
\def\vu{{\bm{u}}}
\def\vv{{\bm{v}}}
\def\vw{{\bm{w}}}
\def\vx{{\bm{x}}}
\def\vtx{\widetilde{\bm{x}}}
\def\vy{{\bm{y}}}
\def\vz{{\bm{z}}}

\def\evalpha{{\alpha}}
\def\evbeta{{\beta}}
\def\evepsilon{{\epsilon}}
\def\evlambda{{\lambda}}
\def\evomega{{\omega}}
\def\evmu{{\mu}}
\def\evpsi{{\psi}}
\def\evsigma{{\sigma}}
\def\evtheta{{\theta}}
\def\eva{{a}}
\def\evb{{b}}
\def\evc{{c}}
\def\evd{{d}}
\def\eve{{e}}
\def\evf{{f}}
\def\evg{{g}}
\def\evh{{h}}
\def\evi{{i}}
\def\evj{{j}}
\def\evk{{k}}
\def\evl{{l}}
\def\evm{{m}}
\def\evn{{n}}
\def\evo{{o}}
\def\evp{{p}}
\def\evq{{q}}
\def\evr{{r}}
\def\evs{{s}}
\def\evt{{t}}
\def\evu{{u}}
\def\evv{{v}}
\def\evw{{w}}
\def\evx{{x}}
\def\evy{{y}}
\def\evz{{z}}

\def\mA{{\bm{A}}}
\def\mB{{\bm{B}}}
\def\mC{{\bm{C}}}
\def\mD{{\bm{D}}}
\def\mE{{\bm{E}}}
\def\mF{{\bm{F}}}
\def\mG{{\bm{G}}}
\def\mH{{\bm{H}}}
\def\mI{{\bm{I}}}
\def\mJ{{\bm{J}}}
\def\mK{{\bm{K}}}
\def\mL{{\bm{L}}}
\def\mM{{\bm{M}}}
\def\mN{{\bm{N}}}
\def\mO{{\bm{O}}}
\def\mP{{\bm{P}}}
\def\mQ{{\bm{Q}}}
\def\mR{{\bm{R}}}
\def\mS{{\bm{S}}}
\def\mT{{\bm{T}}}
\def\mU{{\bm{U}}}
\def\mV{{\bm{V}}}
\def\mW{{\bm{W}}}
\def\mX{{\bm{X}}}
\def\mY{{\bm{Y}}}
\def\mZ{{\bm{Z}}}
\def\mBeta{{\bm{\beta}}}
\def\mPhi{{\bm{\Phi}}}
\def\mLambda{{\bm{\Lambda}}}
\def\mSigma{{\bm{\Sigma}}}

\newcommand{\tens}[1]{\bm{\mathsfit{#1}}}
\def\tA{{\tens{A}}}
\def\tB{{\tens{B}}}
\def\tC{{\tens{C}}}
\def\tD{{\tens{D}}}
\def\tE{{\tens{E}}}
\def\tF{{\tens{F}}}
\def\tG{{\tens{G}}}
\def\tH{{\tens{H}}}
\def\tI{{\tens{I}}}
\def\tJ{{\tens{J}}}
\def\tK{{\tens{K}}}
\def\tL{{\tens{L}}}
\def\tM{{\tens{M}}}
\def\tN{{\tens{N}}}
\def\tO{{\tens{O}}}
\def\tP{{\tens{P}}}
\def\tQ{{\tens{Q}}}
\def\tR{{\tens{R}}}
\def\tS{{\tens{S}}}
\def\tT{{\tens{T}}}
\def\tU{{\tens{U}}}
\def\tV{{\tens{V}}}
\def\tW{{\tens{W}}}
\def\tX{{\tens{X}}}
\def\tY{{\tens{Y}}}
\def\tZ{{\tens{Z}}}

\def\gA{{\mathcal{A}}}
\def\gB{{\mathcal{B}}}
\def\gC{{\mathcal{C}}}
\def\gD{{\mathcal{D}}}
\def\gE{{\mathcal{E}}}
\def\gF{{\mathcal{F}}}
\def\gG{{\mathcal{G}}}
\def\gH{{\mathcal{H}}}
\def\gI{{\mathcal{I}}}
\def\gJ{{\mathcal{J}}}
\def\gK{{\mathcal{K}}}
\def\gL{{\mathcal{L}}}
\def\gM{{\mathcal{M}}}
\def\gN{{\mathcal{N}}}
\def\gO{{\mathcal{O}}}
\def\gP{{\mathcal{P}}}
\def\gQ{{\mathcal{Q}}}
\def\gR{{\mathcal{R}}}
\def\gS{{\mathcal{S}}}
\def\gT{{\mathcal{T}}}
\def\gU{{\mathcal{U}}}
\def\gV{{\mathcal{V}}}
\def\gW{{\mathcal{W}}}
\def\gX{{\mathcal{X}}}
\def\gY{{\mathcal{Y}}}
\def\gZ{{\mathcal{Z}}}

\def\sA{{\mathbb{A}}}
\def\sB{{\mathbb{B}}}
\def\sC{{\mathbb{C}}}
\def\sD{{\mathbb{D}}}
\def\sF{{\mathbb{F}}}
\def\sG{{\mathbb{G}}}
\def\sH{{\mathbb{H}}}
\def\sI{{\mathbb{I}}}
\def\sJ{{\mathbb{J}}}
\def\sK{{\mathbb{K}}}
\def\sL{{\mathbb{L}}}
\def\sM{{\mathbb{M}}}
\def\sN{{\mathbb{N}}}
\def\sO{{\mathbb{O}}}
\def\sP{{\mathbb{P}}}
\def\sQ{{\mathbb{Q}}}
\def\sR{{\mathbb{R}}}
\def\sS{{\mathbb{S}}}
\def\sT{{\mathbb{T}}}
\def\sU{{\mathbb{U}}}
\def\sV{{\mathbb{V}}}
\def\sW{{\mathbb{W}}}
\def\sX{{\mathbb{X}}}
\def\sY{{\mathbb{Y}}}
\def\sZ{{\mathbb{Z}}}

\def\emLambda{{\Lambda}}
\def\emA{{A}}
\def\emB{{B}}
\def\emC{{C}}
\def\emD{{D}}
\def\emE{{E}}
\def\emF{{F}}
\def\emG{{G}}
\def\emH{{H}}
\def\emI{{I}}
\def\emJ{{J}}
\def\emK{{K}}
\def\emL{{L}}
\def\emM{{M}}
\def\emN{{N}}
\def\emO{{O}}
\def\emP{{P}}
\def\emQ{{Q}}
\def\emR{{R}}
\def\emS{{S}}
\def\emT{{T}}
\def\emU{{U}}
\def\emV{{V}}
\def\emW{{W}}
\def\emX{{X}}
\def\emY{{Y}}
\def\emZ{{Z}}
\def\emSigma{{\Sigma}}

\newcommand{\etens}[1]{\mathsfit{#1}}
\def\etLambda{{\etens{\Lambda}}}
\def\etA{{\etens{A}}}
\def\etB{{\etens{B}}}
\def\etC{{\etens{C}}}
\def\etD{{\etens{D}}}
\def\etE{{\etens{E}}}
\def\etF{{\etens{F}}}
\def\etG{{\etens{G}}}
\def\etH{{\etens{H}}}
\def\etI{{\etens{I}}}
\def\etJ{{\etens{J}}}
\def\etK{{\etens{K}}}
\def\etL{{\etens{L}}}
\def\etM{{\etens{M}}}
\def\etN{{\etens{N}}}
\def\etO{{\etens{O}}}
\def\etP{{\etens{P}}}
\def\etQ{{\etens{Q}}}
\def\etR{{\etens{R}}}
\def\etS{{\etens{S}}}
\def\etT{{\etens{T}}}
\def\etU{{\etens{U}}}
\def\etV{{\etens{V}}}
\def\etW{{\etens{W}}}
\def\etX{{\etens{X}}}
\def\etY{{\etens{Y}}}
\def\etZ{{\etens{Z}}}

\newcommand{\pdata}{p_{\rm{data}}}
\newcommand{\ptrain}{\hat{p}_{\rm{data}}}
\newcommand{\Ptrain}{\hat{P}_{\rm{data}}}
\newcommand{\pmodel}{p_{\rm{model}}}
\newcommand{\Pmodel}{P_{\rm{model}}}
\newcommand{\ptildemodel}{\tilde{p}_{\rm{model}}}
\newcommand{\pencode}{p_{\rm{encoder}}}
\newcommand{\pdecode}{p_{\rm{decoder}}}
\newcommand{\precons}{p_{\rm{reconstruct}}}
\newcommand{\one}[1]{\mathbbm{1}{[#1]}}
\newcommand{\laplace}{\mathrm{Laplace}} 

\newcommand{\E}{\mathbb{E}}
\newcommand{\Ls}{\mathcal{L}}
\newcommand{\R}{\mathbb{R}}
\newcommand{\emp}{\tilde{p}}
\newcommand{\lr}{\alpha}
\newcommand{\reg}{\lambda}
\newcommand{\rect}{\mathrm{rectifier}}
\newcommand{\softmax}{\mathrm{softmax}}
\newcommand{\sigmoid}{\sigma}
\newcommand{\softplus}{\zeta}
\newcommand{\KL}{D_{\mathrm{KL}}}
\newcommand{\Var}{\mathrm{Var}}
\newcommand{\standarderror}{\mathrm{SE}}
\newcommand{\Cov}{\mathrm{Cov}}
\newcommand{\normlzero}{L^0}
\newcommand{\normlone}{L^1}
\newcommand{\normltwo}{L^2}
\newcommand{\normlp}{L^p}
\newcommand{\normmax}{L^\infty}

\newcommand{\parents}{Pa} 

\section{Introduction}
In the past few years, owing to the success of Transformer-based~\cite{vaswani2017attention} pre-trained models, such as BERT~\cite{devlin2019bert}, RoBERTa\cite{liu2019roberta}, GPT2\cite{radford2019language}, \etc we have experienced a performance breakthrough in natural language processing (NLP) tasks. With a small amount of fine-tuning, the pre-trained models can achieve state-of-the-art performance across different tasks~\cite{devlin2019bert,liu2019roberta, radford2019language}. Nevertheless, the outperforming models are evaluated in offline settings, and the inference latency is not assessed or considered as a quality factor.

However, adapting and deploying such large pre-trained models to production systems (\eg online shopping services) is not straightforward due to the latency constraint and the large volume of incoming requests (\eg millions of requests per second). Prior to this work, researchers have proposed to compress a large model via either model pruning~\cite{michel2019sixteen,fan2019reducing,hou2020dynabert} or token pruning~\cite{wang2021spatten,goyal2020power,kim2021learned}. In addition, compressing a large teacher model into a compact model via knowledge distillation has been studied extensively in the past ~\cite{sanh2019distilbert,sun2019patient, sun2020mobilebert,jiao2020tinybert, he2021generate}. Finally, another line of work targets on plugging multiple sub-classifiers into deep neural networks to enable a flexible computation on demand, \textit{a.k.a.,} early exiting~\cite{7900006, pmlr-v97-kaya19a,schwartz2020right, liu2020fastbert}

\begin{figure}
    \centering
    \vspace{-5mm}
    \begin{subfigure}[t]{0.4\textwidth}
    \centering
     \scalebox{0.25}{
    \includegraphics{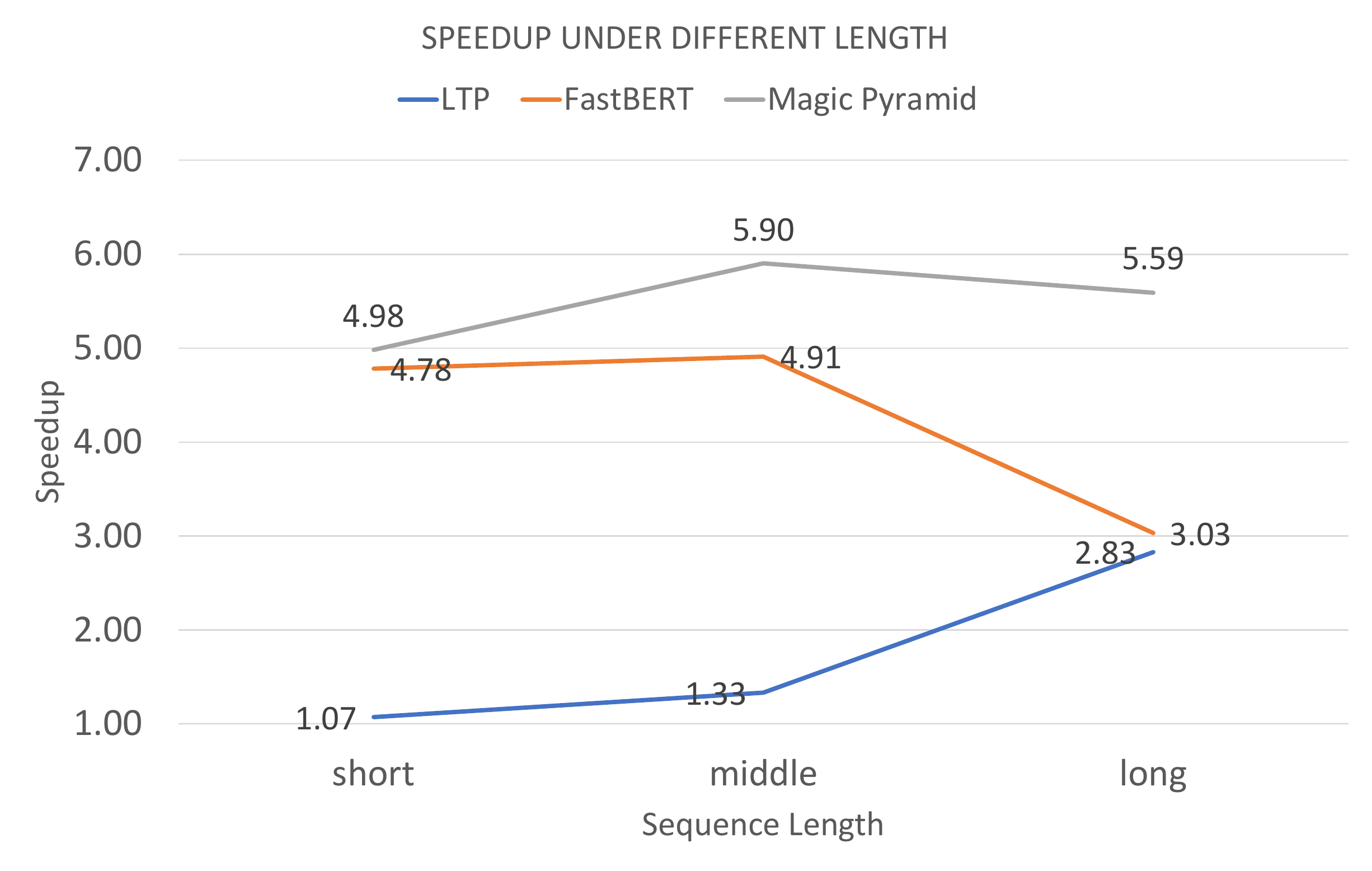}
     }
    \caption{Yelp.}
    \end{subfigure}
    \hspace{15mm}
    \begin{subfigure}[t]{0.4\textwidth}
    \centering
     \scalebox{0.25}{
    \includegraphics{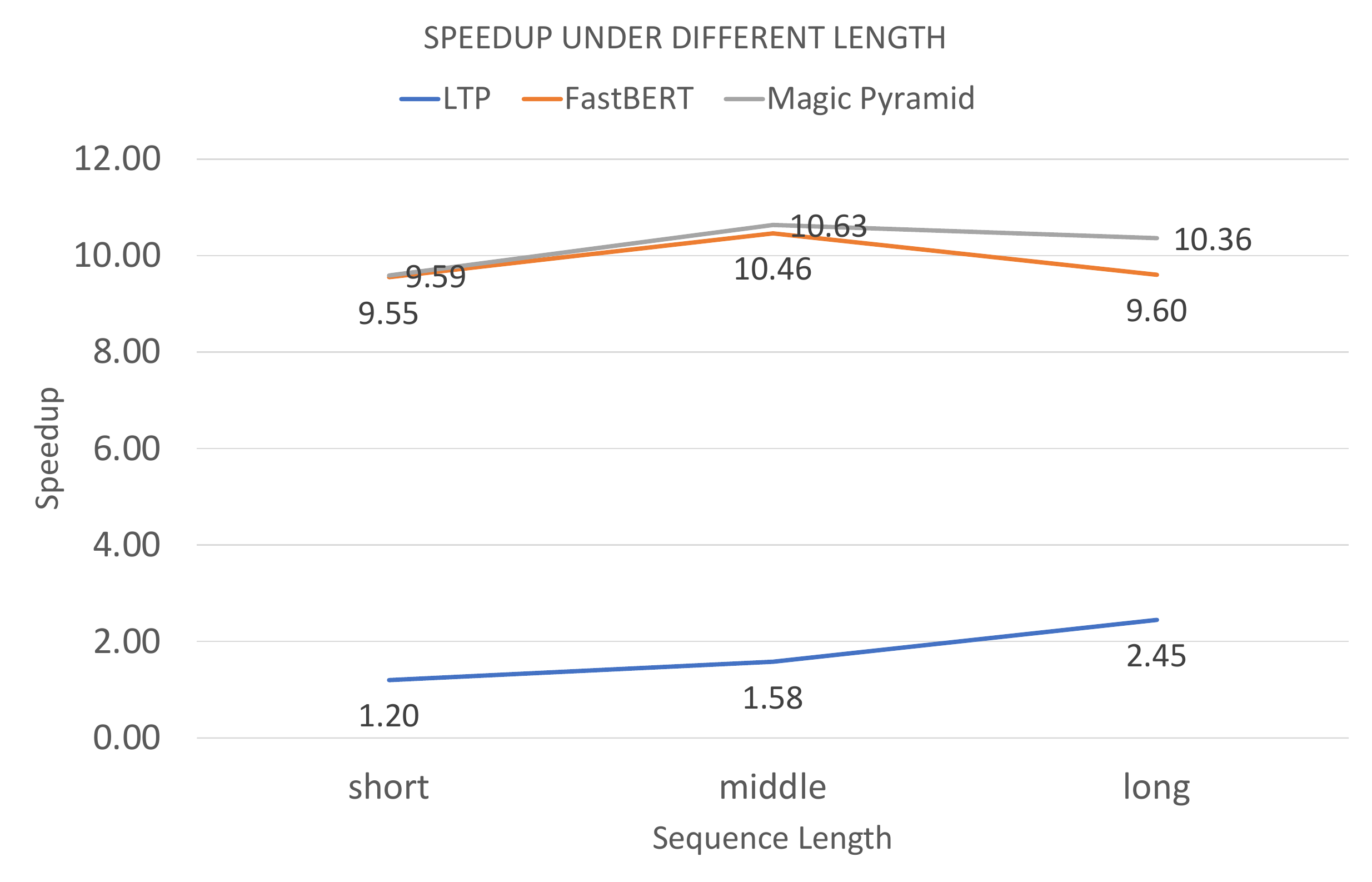}
     }
    \caption{AG news}
    \end{subfigure}
   \caption{Speedup of LTP (token pruning), FastBERT (early exiting) and MP (ours) under different sequence lengths on Yelp and AG news. X axis is sequence length, while Y axis is speedup. short: 1-35 tokens; middle: 35-70 tokens; long: \textgreater 70 tokens}
  \label{fig:diff_len}
  \vspace{-4mm}
\end{figure}

The token pruning concentrates on a width-wise computational reduction,  whereas the early exiting succeeds in a depth-wise inference acceleration. Our study shows that for certain tasks where the input data is diverse (in terms of sequence length), these two latency reduction methods perform in the opposite direction. As illustrated in \figref{fig:diff_len} (a), speedup (Y axis) achieved via early exiting (FastBERT) decreases for long input sizes (X axis). However, token pruning (LTP) speedup rises as the input size increases. We believe these two approaches are orthogonal and can be combined into a single model to maintain the latency reduction gain across the variable input length. In this work, we present a novel approach, \textit{Magic Pyramid} (MP), to encourage a speed-adjustable inference. The contribution of this paper includes:
\begin{itemize}
    \item Our empirical study shows that token pruning and early exiting are potentially orthogonal. This motivates further research on employing the two orthogonal inference optimization methods  within a single model for model inference acceleration.
    \item We propose a method (referred to as Magic Pyramid) to exploit the synergy between token pruning and early exiting and attain higher computational reduction from width and depth perspectives. 
    \item Compared to two strong baselines, our approach can significantly accelerate the inference time with an additional 0.5-2x speedup but less than 0.5\% degradation on accuracy across five classification tasks.
\end{itemize}

\section{Related Work}
\if
Learned Token Pruning for Transformers

SpAtten: Efficient Sparse Attention Architecture with Cascade Token and Head Pruning

PoWER-BERT: Accelerating BERT Inference via Progressive Word-vector Elimination

Reviewing missing parts in these papers
\fi 

Large pre-trained models have demonstrated that increasing the model capacity can pave the way for the development of superior AI. However, as we have limited resources allocated for production systems, there has been a surge of interest in efficient inference. Previous works~\cite{sanh2019distilbert,sun2019patient, sun2020mobilebert,jiao2020tinybert, he2021generate} have opened a window into an effective model compression via knowledge distillation (KD)~\cite{hinton2015distilling}. The core of KD is to use a compact student model to mimic the behavior or structure of a large teacher model. As such, the performance of the student model is as accurate as its teacher, but consuming less computation.

Other researchers approach efficient inference by manipulating the original model. One elegant solution is pruning, which can reduce the computation by removing non-essential components. These components can be either model parameters (model pruning)~\cite{michel2019sixteen,fan2019reducing,hou2020dynabert} or tokens (token pruning)~\cite{wang2021spatten,goyal2020power,kim2021learned}. In addition, one can boost the speed of the numerical operations of a model through quantization~\cite{zafrirq8bert,wrobel2020compression,shen2020q}.

The aforementioned works lack flexibility in terms of the speedup, albeit some success. To satisfy varying demands, we have to train multiple models. Since deep neural networks can be considered as a stack of basic building blocks, a list of works introduces early exiting~\cite{7900006, pmlr-v97-kaya19a,schwartz2020right, liu2020fastbert}, which attaches a set of sub-classifiers to these sub-networks to encourage an adjustable inference within a single model, when needed. As opposed to the prior works, which focus on the one-dimensional speedup, this work takes the first step to superimpose token pruning on early exiting. Our empirical studies confirm that these two approaches can accelerate the inference collaboratively and significantly.

\section{Methodology - Proposed Method}
\label{sec:method}
\begin{figure}[h]
    \centering
    \scalebox{0.55}{
    \includegraphics[width=\textwidth]{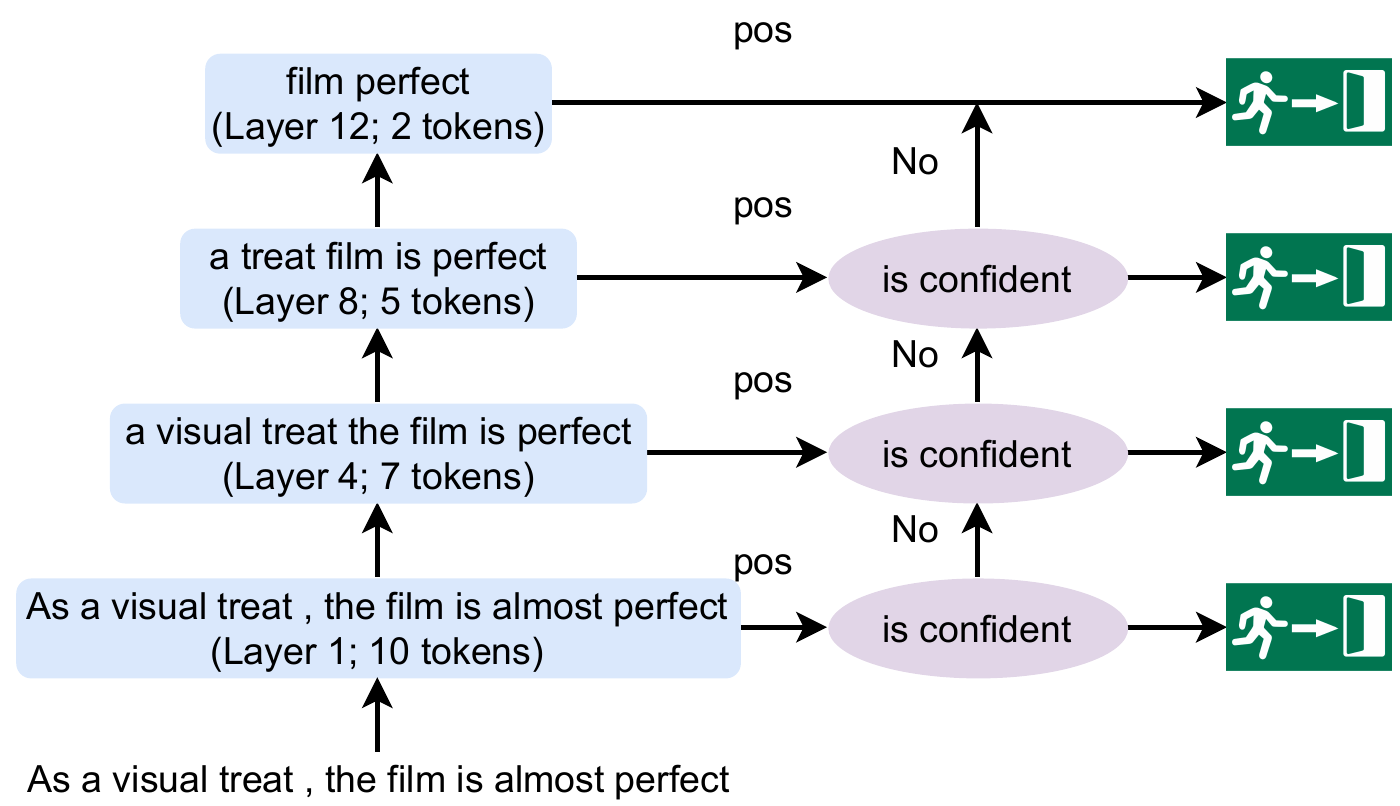}
    }
    \caption{Schematic illustration of magic pyramid for a sentiment analysis task}
    \label{fig:mp}
    \vspace{-3mm}
\end{figure}

Prior to this work, token pruning and early exiting have been proven to be effective in accelerating the inference~~\cite{wang2021spatten,goyal2020power,schwartz2020right, liu2020fastbert,kim2021learned}. However, as shown in \figref{fig:diff_len}, these approaches fall short of reducing the latency at two ends, \ie short sequences and long sequences. For example, \figref{fig:diff_len} (a) shows that LTP (token-pruning) provides the highest speed-up for long input sequences. While FastBERT (early exiting) speedup drops as the input size increases from short to long. Therefore we propose a novel approach: \textit{Magic Pyramid} (MP), which benefits from a combination of token pruning and early exiting. \figref{fig:mp} provides a schematic illustration of MP. First of all, MP enables to terminate an inference at any layer when needed. Second, with the increase of the depth of Transformer, redundant tokens can be expelled. The detailed designs are provided in the rest of this section.

\paragraph{Transformer architecture} Owing to it outstanding performance, Transformer~\cite{vaswani2017attention} has become a de facto model for NLP tasks, especially after the triumph of pre-trained language models~\cite{devlin2019bert,liu2019roberta,radford2019language}. A standard Transformer is comprised of $L$ stacked Transformer blocks: $\{\mathrm{B}_l\}_{l=1}^L$, where each block $\mathrm{B}_l$ is formulated as:
\begin{align}
    H'_l  &=   \mathrm{MultiHead}(H_{l-1}) \\
    H_l & =  \mathrm{FFN}(H'_l)
\end{align}
where $H\in\mathbb{R}^{n\times d}$ and $H'\in\mathbb{R}^{n\times d}$ are the hidden states. $n$ is sequence length, while $d$ is the feature dimension. $\mathrm{MultiHead}$ and $\mathrm{FFN}$ are multi-head attention module and position-wise feedforward module respectively. We omit $\mathrm{Residual}$ module and $\mathrm{Layernorm}$ module in between for simplicity.


\paragraph{Early exiting} As shown in \figref{fig:mp}, in addition to the Transformer backbone and a main classifier, one has to attach an individual sub-classifier module ($\mathrm{subclassifier}(\cdot)$) to the $\mathrm{FFN}$ of each Transformer block $\mathrm{B}_i$. As such, one can choose to terminate the computation at any layer, when a halt value $\tau$ is reached.

Following~\cite{liu2020fastbert}, the $\mathrm{subclassifier}(\cdot)$ consists of a Transformer block $B$, a pooling layer $\mathrm{Pooler}$ and a projection layer $\mathrm{Projector}$ with a $\mathrm{softmax}$ function. $\mathrm{Pooler}$ extracts the hidden states of $\mathrm{[CLS]}$ as the representation of the input, while $\mathrm{Projector}$ projects the dense vector into $N$-class logits.

Similar to ~\cite{liu2020fastbert}, we leverage a two-stage fine-tuning to enhance the performance of sub-classifiers via knowledge distillation. Specifically, we first train the Transformer backbone and the primary classifier through a standard cross entropy between the ground truth $y$ and the predictions $y'$. Afterward, we freeze the backbone and the primary classifier, but train each $\mathrm{subclassifier}(\cdot)$ via a Kullback–Leibler divergence:
\begin{align}
    D_{KL}(p_s, p_t)=\sum_{i=1}^N p_s(i) * \log \frac{p_s(i)}{p_t(i)}
\end{align}
where $p_s$ and $p_t$ are the predicted probability distribution from the $\mathrm{subclassifier}(\cdot)$ and the main classifier respectively. Since there are $L-1$ sub-classifiers, the loss of the second stage can be formulated as:
\begin{align}
   \mathcal{L}(p_{s_1},...,p_{s_{L-1}}, p_t) = \sum_{l=1}^{L-1}D_{KL}(p_{s_i}, p_t)
   \label{equ:kd}
\end{align}

Once all modules are well-trained, we can stitch them together to achieve a speed-adjustable inference. At each layer $l$, we first obtain the hidden states $H_l$ from the Transformer block $B_l$. Then a probability $p_{s_l}$ can be computed from $\mathrm{subclassifier}(H_l)$. One can use $p_{s_l}$ to calculate the uncertainty $u_l$ via:
\begin{align}
   u_l= \frac{\sum p_{s_l}\log p_{s_l}}{\log \frac{1}{N}}
\end{align}
where $u_l$ is bound to $\{0,1\}$. If $u_l \leq \tau$, we can terminate the computation. A larger $\tau$ suggests a faster exit.

\begin{figure}[h]
    \centering
    \scalebox{0.9}{
    \includegraphics[width=0.45\textwidth]{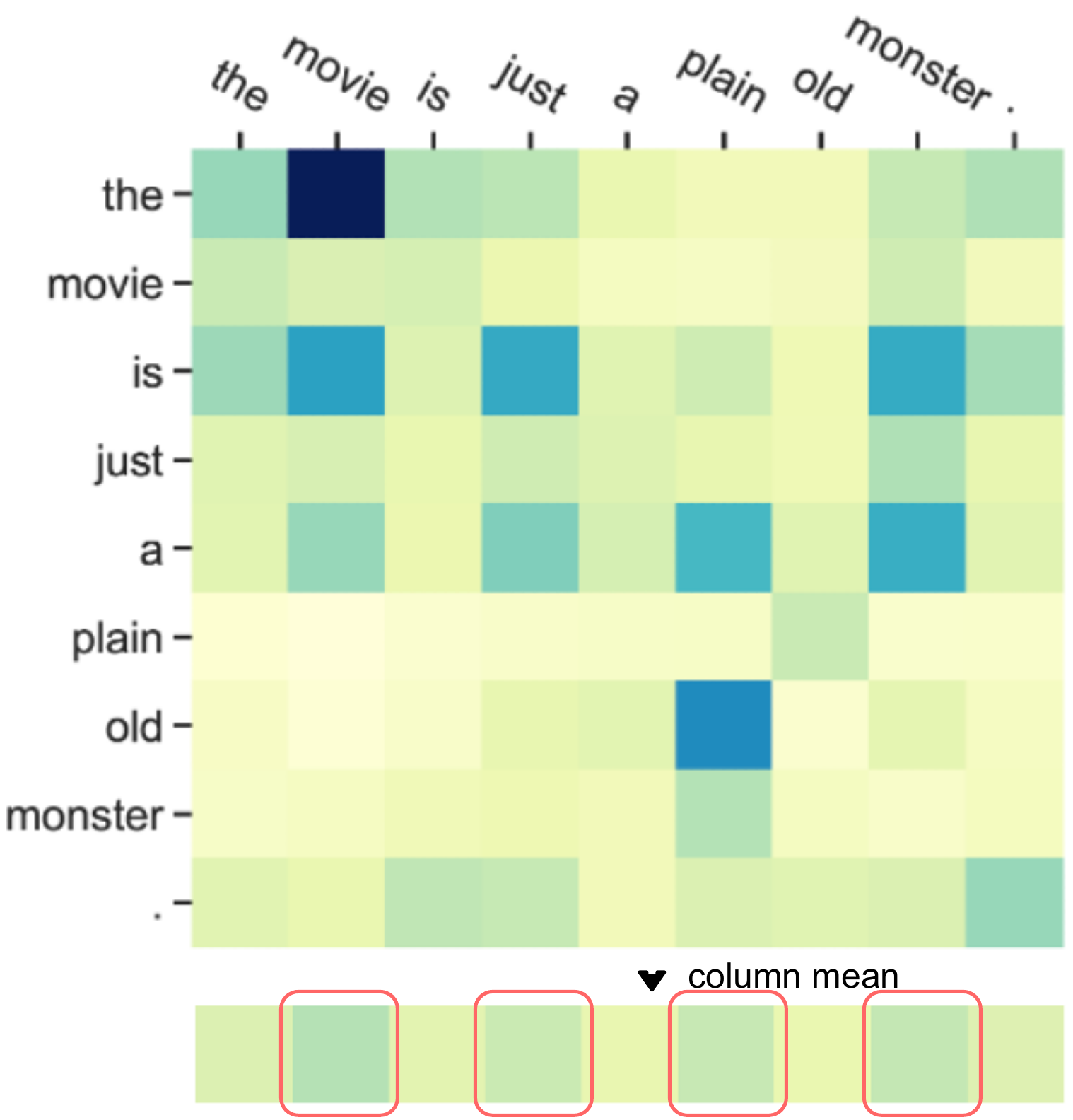}
    }
    \caption{An example of attention probability in a single head. Darker color suggests a higher attention score. The bottom heatmap is the column mean of the attention matrix.}
    \label{fig:attn}
    \vspace{-5mm}
\end{figure}

\paragraph{Token pruning} The core of the Transformer block is the $\mathrm{MultiHead}$ module, which is responsible for a context-aware encoding of each token. Notably, we compute pairwise importance among all tokens within the input via self-attention. The attention score of each head $h$ between $x_i$ and $x_j$ is obtained from:
\begin{align}
   A^h(x_i, x_j) = \mathrm{softmax}(\frac{H(x_i)^T\rmW^T_q\rmW_kH(x_j)}{\sqrt{d}})
\end{align}
where $H(x_i)\in\mathbb{R}^d$ and $H(x_j)\in\mathbb{R}^d$ are the hidden states of $x_i$ and $x_j$ respectively. $\rmW_q\in \mathbb{R}^{d_h\times d}$ and $\rmW_k\in \mathbb{R}^{d_h\times d}$ are learnable parameters. $d_h$ is set to $d/N_h$, and $N_h$ is the number of heads used in the Transformer block $B$.
Since we have to conduct $n^2$ such operation to acquire an attention score matrix $A\in\mathbb{R}^{n\times n}$, the complexity of $\mathrm{MultiHead}$ quadratically scales with the sequence length. Therefore, we encounter a computational bottleneck when working on long sequences. However, if we take the average of $A$ along the $i$th column, we notice that the different tokens have distinctive scores as shown in \figref{fig:attn}. Tokens with large scores tend to be more salient than others, as they receive more attention. As such, we can prune the non-salient tokens to save the computation. We formally define an importance score of each token $x_i$ at layer $l$ as:
\begin{align}
   s^l(x_i) = \frac{1}{N_h}\frac{1}{n}\sum_{h=1}^{N_h}\sum_{j=1}^n A^h_l(x_i, x_j)
\end{align}

Before this work, researchers have proposed two approaches to remove the unimportant tokens based on $s^l(\cdot)$: \textit{i)} top-k based pruning and \textit{ii)} threshold-based pruning~\cite{goyal2020power, wang2021spatten, kim2021learned}. In this work, we follow~\cite{kim2021learned}, which leverages a layer-wise learnable threshold to achieve a fast inference and is superior to other works\cite{wang2021spatten,goyal2020power}. We first fine-tune Transformer parameters $\Theta$ on a downstream task. Then we introduce a two-stage pruning scheme to seek a suitable threshold $\Delta$ and $\Theta$, which can accelerate the inference and maintain a decent accuracy. In the first pruning stage, we apply a gating function $\sigma(\cdot)$ to weight the outputs $H_l$ from the current layer $l$, before we pass them to the next layer $l+1$ like this:
\begin{align}
   M^l(x_i) & =  \sigma(\frac{s^l(x_i)-\Delta^l}{T}) \\
   \hat{H}_l(x_i) & = H_l(x_i) \otimes M^l(x_i)
\end{align}
where $\sigma$ is a sigmoid function, $\otimes$ is an element-wise multiplication, $T$ is a temperature parameter and $\Delta^l\in \mathbb{R}$ is a learnable threshold at layer $l$. If $M^l(x_i)$ approaches zero, $\hat{H}_l(x_i)$ will become zero as well. As such, $\hat{H}_l(x_i)$ has no impact on the subsequent layers. At this stage, since $\sigma(\cdot)$ allows the flow of the back-propagation, both $\Theta$ and $\Delta$ can be optimized. In addition, \cite{kim2021learned} also impose a $L1$ loss on $M$ as a regularizer to encourage the pruning operation. Please refer to their paper for the details.

In the second pruning stage, we binarize the mask values at the inference time via:
\begin{equation}
    M^l(x_i) = 
    \begin{dcases}
    1,           & s^l(x_i)-\Delta^l>0.5 \\
    0,              & \text{otherwise}
    \end{dcases}
\end{equation}
If $s^l(x_i)$ is below the threshold $\Delta^l$, $x_i$ is subject to the removal from layer $l$ and will not contribute towards the final predictions. We freeze $\Delta$ but update $\Theta$, such that the model can learn to accurately predict the labels merely conditioning on the retained tokens.



\section{Experiments}

To examine the effectiveness of the proposed approach, we use five language understanding tasks as the testbed.  We describe our datasets and our experimental setup in the following.
\begin{table}[h]
    \centering
    \begin{tabular}{lccl}
    \toprule
        Data & Train & Test & Task\\
        \midrule
        AG news &  120K  &7.6K & topic\\
        Yelp & 560K& 38K& sentiment \\
         QQP & 364K & 40K & paraphrase  \\
        MRPC & 3.7K & 408 & paraphrase \\
        RTE & 2.5K & 277 & language inference \\
        \bottomrule
    \end{tabular}
    \caption{The statistics of datasets}
    \vspace{-9mm}
    \label{tab:data}
\end{table}

\paragraph{Datasets} The first two tasks are: \textit{i)} AG news topic identification~\cite{zhang2015character}, and \textit{ii)} Yelp polarity sentiment classification~\cite{zhang2015character}. The last three are: \textit{i)} Quora Question Pairs (QQP) similarity detection dataset, \textit{ii)} Microsoft Research Paraphrase Corpus (MRPC) dataset and \textit{iii)} Recognizing Textual Entailment (RTE) dataset. All datasets are from GLUE benchmark~\cite{wang2018glue} and focus on predicting  between a sentence pair. The datasets are summarized in \tabref{tab:data}.
\begin{table}[h]
    \centering
    \scalebox{0.9}{
    \begin{tabular}{c|ccccc}
    \toprule
        Datasets & BERT & DistilBERT & LTP & FastBERT & MP (ours)\\
        \midrule
       AG news & 3,-,-,- & 3,-,-,- & 3,1,2,- & 3,-,-,2 & 3,1,2,2\\
       Yelp &3,-,-,-  & 3,-,-,- & 3,1,2,- & 3,-,-,2 & 3,1,2,2\\
        QQP  &5,-,-,-, & 5,-,-,-, & 5,2,5,- & 5,-,-,5 & 5,2,5,5\\
       MRPC &10,-,-,-,  & 10,-,-,-, & 10,10,5,- & 10,-,-,5 & 10,10,5,5\\
       RTE & 10,-,-,-,& 10,-,-,-, & 10,10,5,- & 10,-,-,5 & 10,10,5,5\\
       \bottomrule
    \end{tabular}
    }
    \caption{The number of epochs used for regular training, soft pruning, hard pruning, subclassifiers training on different datasets. ``-" indicates the corresponding stage is inactive.}
    \label{tab:training}
    \vspace{-8mm}
\end{table}

\paragraph{Experimental setup} We compare our approach with four baselines: \textit{i)} standard BERT~\cite{devlin2019bert}, \textit{ii)} distilBERT~\cite{sanh2019distilbert}, \textit{iii)} learned token pruning (LTP)~\cite{kim2021learned} and \textit{iv)} FastBERT~\cite{liu2020fastbert}. Except distilBERT, all approaches are fine-tuned on uncased BERT-base model (12 layers).

For training, we use a batch size of 32 for QQP, MRPC, and RTE. We set this to 64 for AG news and Yelp. Since different approaches adopt different training strategies, we unify them as four steps:
\begin{enumerate}
    \item Regular training: training a model $\Theta$ without additional components;
    \item Soft pruning: training a model $\Theta$ and threshold $\Delta$;
    \item Hard pruning: training a model $\Theta$ with the binarized mask values;
    \item Sub-classifiers training: training sub-classifiers on Equ. \eqref{equ:kd}. For MP, we also activate the pruning operations.
\end{enumerate}
We report the number of the training epochs of different steps for all approaches in \tabref{tab:training}. Similar to \cite{kim2021learned}, we vary the threshold of the final layer $\Delta^L$ from 0.01 to 0.08, and the threshold for $\Delta^l$ is set to $\Delta^Ll/L$. We search the temperature $T$ in a search space of \{1e\textminus5,2e\textminus5,5e\textminus5\} and vary $\lambda$ from 0.001 to 0.2. We use a learning rate of 2e\textminus5 for all experiments. We consider accuracy for the classification performance and giga floating point operations (GFLOPs) for the speedup.

\begin{table*}[ht]
    \centering
    \scalebox{0.78}{
    \begin{tabular}{l|rr|rr|rr|rr|rr}
    \toprule
       & \multicolumn{2}{c|}{AG news} & \multicolumn{2}{c|}{Yelp} & \multicolumn{2}{c|}{QQP} & \multicolumn{2}{c|}{MRPC}& \multicolumn{2}{c}{RTE} \\
       & Acc. & GFLOPs & Acc. & GFLOPs & Acc. & GFLOPs & Acc. & GFLOPs & Acc. & GFLOPs \\
       \midrule
       BERT & 94.3  & 9.0 (1.00x) & 95.8 & 17.2 (1.00x) & 91.3 & 5.1 (1.00x) & 85.3 & 9.2 (1.00x)& 68.6 & 11.2 (1.00x)\\
       distilBERT & 94.4 & 4.5 (2.00x) & 95.7 & 8.6 (2.00x) & 90.4 & 2.6 (2.00x) & 84.6 &4.6 (2.00x)& 58.8 & 5.6 (2.00x)\\
       LTP&   94.3 & 5.3 (1.72x)& 94.7 & 7.4 (2.32x) & 90.6 & 3.2 (1.60x)& 84.8 &  6.2 (1.48x)& 67.8& 7.5 (1.50x)\\
      FastBERT& 94.3  &  2.3 (3.97x)&  94.8& 2.8 (6.18x) & 90.7 & 1.6 (3.20x) &  84.3  & 4.3 (2.13x)& 67.6&  8.4 (1.33x)\\
      MP (ours)& 94.3 & 1.8 (4.95x)&94.5 & 2.1 (8.25x) & 90.4 & 1.3 (4.03x) &  83.8 & 3.3 (2.77x) & 67.5 & 6.5 (1.72x) \\
       \bottomrule
    \end{tabular}
    }
    \caption{The accuracy and GFLOPs of BERT~\cite{devlin2019bert}, distilBERT~\cite{sanh2019distilbert}, LTP (learned token pruning)~\cite{kim2021learned}, FastBERT~\cite{liu2020fastbert} and MP (ours) on different datasets. The numbers in parentheses are speedup.}
    \label{tab:main}
    \vspace{-3mm}
\end{table*}

\begin{table}[h]
    \centering
    \scalebox{0.9}{
    \begin{tabular}{c|ccc|ccc}
    \toprule
    & \multicolumn{3}{c|}{AG news} & \multicolumn{3}{c}{Yelp} \\
    \midrule 
    $\tau$ & 0.1 & 0.5 & 0.8 &  0.1 & 0.5 & 0.8  \\
    \midrule
     FastBERT  & 3.97x & 10.30x & 11.95x &  3.15x & 6.18x & \ \ \ 8.84x\\
        MP (ours) & 4.95x & 10.53x & 11.95x & 5.35x& 8.25x & 10.10x\\ 
        \bottomrule
    \end{tabular}
    }
    \caption{Speedup of FastBERT and MP with different $\tau$.}
    \label{tab:speed}
    \vspace{-4mm}
\end{table}

For token pruning approach, previous works~\cite{wang2021spatten,kim2021learned} have shown that there exits a trade-off between accuracy and speedup. Thus, we report the performance of models achieving smallest GFLOPs with at most 1\% accuracy drop compared to the BERT baseline. Similarly, the speedup of FastBERT is also controlled by the halt value $\tau$. We select $\tau$ obtaining a on-par accuracy with the token pruning competitors for the sake of a fair comparison. This selection criterion is applied to MP as well.

\tabref{tab:main} demonstrates that all approaches experience loss in accuracy, when a fast inference is activated. Overall, FastBERT is superior to disilBERT and LTP in terms of both accuracy and GFLOPs. Under the similar accuracy, our approach manages to have a significantly faster inference than FastBERT, which leads to up to 2.13x extra speedup. We notice that the speedup and accuracy also correlate to the complexity of tasks and the number of training data. Specifically, for the sentence-pair classification tasks, since QQP has much more data (\cf \tabref{tab:data}), it achieves 4.03x speedup with a loss of 1\% accuracy. On the contrary, RTE and MRPC obtain at most 2.77x speedup with the same amount of accuracy degradation. Under the same magnitude of the training data, as AG news and Yelp are simpler than QQP, they can gain up to 8.25x speedup after sacrificing 1\% accuracy.

\paragraph{Gains over FastBERT} In \secref{sec:method}, we have claimed that MP can benefit from both token pruning and early exiting. Although this claim is evidenced in \tabref{tab:main}, we are interested in investigating whether such gains consistently hold, when tuning $\tau$ to control the speed of the inference. According to \tabref{tab:speed}, MP can drastically boost the speedup of FastBERT, except for an aggressive $\tau$, which will cause the computation to terminate at the first two layers.

\paragraph{Speedup on sequences with different lengths} Intuitively, longer sentences tend to have more redundant tokens, which can confuse the lower sub-classifiers. Consequently, longer sentences require more computation before reaching a lower uncertainty $u$. We bucket the Yelp and AG news dataset into three categories: \textit{i)} short sequences (1-35 tokens),  \textit{ii)} middle sequences (35-70 tokens) and \textit{iii)} long sequences (\textgreater 70 tokens). \figref{fig:diff_len} indicates that LTP prefers long sequences, while FastBERT favors short sequences. Since MP combines the early exiting with the token pruning, it can significantly accelerate both short and long sequences, compared to the two baselines.


\section{Conclusion}
In this work, we introduce Magic Pyramid, which can maintain a trade-off between speedup and accuracy for BERT-based models. Since MP is powered by two outstanding efficiency-encouraging approaches, it can yield substantially faster inference over the baselines up to additional 2x speedup. We also found that token pruning and early exiting falls to efficiently handle sequences under certain length groups. In contrary, such limitations can be combated by MP, thereby our approach can indiscriminately accelerate inference for every input data (i.e, inference request) regardless of its length.

\bibliography{refs}
\bibliographystyle{plain}

\end{document}